\documentclass{article}

\usepackage{arxiv}

\usepackage[utf8]{inputenc} 
\usepackage[T1]{fontenc}    
\usepackage{hyperref}       
\usepackage{url}            
\usepackage{booktabs}       
\usepackage{amsfonts}       
\usepackage{nicefrac}       
\usepackage{microtype}      
\usepackage{lipsum}
\usepackage{graphicx}
\usepackage[numbers]{natbib}
\usepackage{amsmath}
\usepackage{listings}
\title{REM-CTX: Automated Peer Review via Reinforcement Learning with Auxiliary Context}

\author{
 Pawin Taechoyotin \\
  Department of Computer Science\\
  University of Colorado Boulder\\
  Boulder, CO 80309 \\
  \texttt{pawin.taechoyotin@colorado.edu} \\
   \And
 Daniel Acuna \\
  Department of Computer Science\\
  University of Colorado Boulder\\
  Boulder, CO 80309 \\
  \texttt{daniel.acuna@colorado.edu} \\
}

\begin{document}
\maketitle
\begin{abstract}
Most automated peer review systems rely on textual manuscript content alone, leaving visual elements such as figures and external scholarly signals underutilized. We introduce REM-CTX, a reinforcement-learning system that incorporates auxiliary context into the review generation process via correspondence-aware reward functions. REM-CTX trains an 8B-parameter language model with Group Relative Policy Optimization (GRPO) and combines a multi-aspect quality reward with two correspondence rewards that explicitly encourage alignment with auxiliary context. Experiments on manuscripts across Computer, Biological, and Physical Sciences show that REM-CTX achieves the highest overall review quality among six baselines, outperforming other systems with substantially larger commercial models, and surpassing the next-best RL baseline across both quality and contextual grounding metrics. Ablation studies confirm that the two correspondence rewards are complementary: each selectively improves its targeted correspondence reward while preserving all quality dimensions, and the full model outperforms all partial variants. Analysis of training dynamics reveals that the criticism aspect is negatively correlated with other metrics during training, suggesting that future studies should group multi-dimension rewards for review generation.
\end{abstract}


\section{Introduction}
Recent research shows how Large Language Models (LLMs) are increasingly integrated across the full scientific workflow, including research ideation, experimentation, review generation, and iterative refinement \citep{lu2024aiscientistfullyautomated, yamada2025aiscientistv2workshoplevelautomated}. With the rapid advancement of LLM capabilities, research on automated peer-review generation has gained increasing attention \citep{lin2023automated,liang2024useful}. These systems range from relatively simple prompting-based approaches that directly generate reviews from manuscript text \citep{bartoli2016paper,kang2018dataset}, to more complex agentic frameworks involving multiple collaborating models \citep{darcy2024marg}, and systems that incorporate external knowledge or multi-modal context \citep{taechoyotin2024mamorx}. Empirical evaluations suggest that such automated systems can produce feedback comparable to human reviewers in certain aspects and occasionally surpass human reviews in consistency and coverage \citep{liang2024useful,taechoyotin2024mamorx}.

Despite these advances, most automated peer-review generation systems rely primarily on internal model knowledge or textual manuscript content alone. This limitation can lead to inaccurate novelty assessments, insufficient grounding in prior literature, and incomplete discussion of visual elements such as figures. One approach to address these gaps is to use multi-modal LLMs, which can directly process visual inputs and provide feedback on figures \citep{taechoyotin2024mamorx}. However, current multi-modal models still have incomplete modality coverage, with no support for (citation) graphs or other non-visual data types. An alternative is to augment a text-based LLM with auxiliary contextual information encoded in text form, and train the model to incorporate it into its outputs.

\begin{figure}[t]
\begin{center}
\includegraphics[width=1\textwidth]{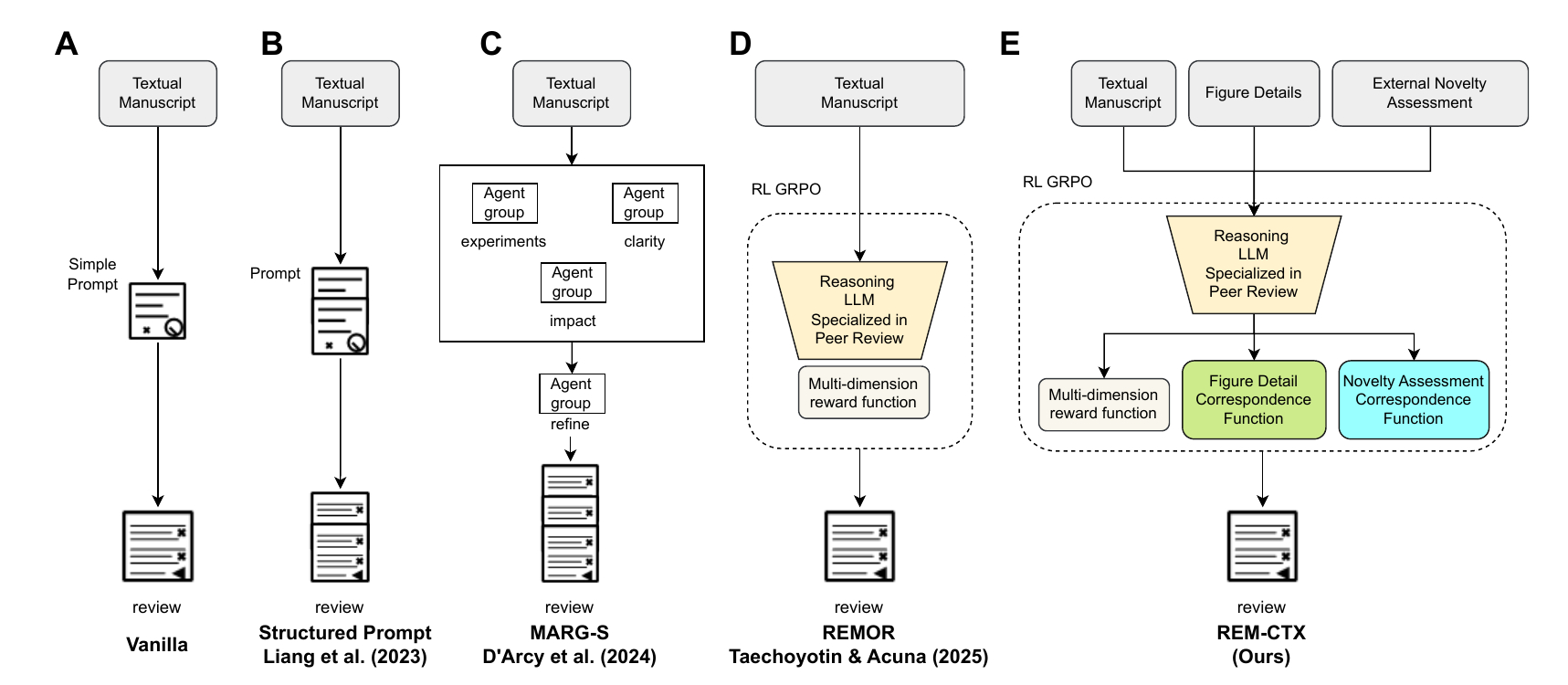}
\end{center}
\caption{Comparison of scientific review generation models. (a) Vanilla or simple prompting-based review generation models, which rely primarily on textual manuscript content and internal model knowledge. (b) Structured, prompting-based review generation model \citep{liang2024useful}. (c) A multi-agent review generation model that incorporates agents to analyze the manuscript from different aspects \citep{darcy2024marg}. (d) REMOR: A reinforcement learning-based review generation model that optimizes review quality using reward functions based the manuscript text only \citep{taechoyotin2025remor}. (e) Our proposed model, REM-CTX, which combines auxiliary context with reinforcement-learning optimization via GRPO and correspondence-aware reward functions, produces more grounded and informative peer reviews.} 
\label{fig:overview}
\end{figure}

In this article, we introduce \textbf{RE}inforcement learning \textbf{M}ulti-objective review generation with auxiliary \textbf{C}on\textbf{T}e\textbf{X}t (REM-CTX) (Figure \ref{fig:overview}), a system that adds auxiliary contexts based on visual components and external knowledge to the review generation input. These contexts were produced by commercial multi-modal large language models for figures and external scholarly datasets and LLMs for novelty assessments. We optimize the model using reinforcement learning to explicitly encourage the addition of these signals into the generated reviews. By incorporating these additional signals, REM-CTX can generate reviews that are more grounded in the manuscript's visual content and the broader scholarly context. Importantly, these auxiliary contexts can include any external information as long as they can be represented as text, a topic for future work.

Our contributions are as follows:
\begin{enumerate}
    \item We propose \textit{correspondence reward functions}, a general mechanism for incentivizing RL-trained language models to incorporate auxiliary contextual information, instantiated here for figure details and novelty assessments.
    
    \item We curate three new datasets: \textit{PeerRTEx}, a multi-domain peer review dataset spanning Computer, Biological, Physical Sciences; \textit{FCRDat}, a sentence-level figure correspondence dataset; and \textit{NCRDat}, a sentence-level novelty correspondence dataset.
    
    \item We demonstrate that REM-CTX achieves the highest overall review quality among six baselines, that the correspondence rewards are complementary and do not degrade quality dimensions, and that training dynamics reveal interpretable trade-offs between review dimensions.
\end{enumerate}

\section{Related Work}
\paragraph{Automated Peer Review Generation} The increasing volume of submissions to scientific venues has motivated research into automated assistance for peer review. Early work explored the feasibility of doing this, demonstrating that structured feedback can be synthesized from manuscript content \citep{bartoli2016paper}. More recently, datasets such as \textit{PeerRead} have enabled the systematic study of peer review generation and analysis in NLP research \citep{kang2018dataset}. Recent advances in Large Language Models (LLMs) have significantly expanded the scope of automated scholarly feedback. Automated Scholarly Paper Review (ASPR) frameworks formalize the task of machine-generated academic reviewing and highlight challenges, including faithfulness, quality evaluation, and ethical considerations \citep{lin2023automated}. Empirical studies further suggest that LLM-generated reviews can provide useful feedback that is comparable in some dimensions to that of human reviewers, although reliability and bias remain concerns \citep{liang2024useful}. AI-assisted peer review workflows have also been proposed to augment, rather than replace, human reviewers, thereby improving efficiency while maintaining oversight \citep{checco2021ai}. Beyond single-model approaches, recent proposals incorporate multi-agent collaboration and multi-modal signals. Multi-agent review generation frameworks simulate multiple reviewers with specialized roles to improve coverage and critique diversity \citep{darcy2024marg}. Similarly, multi-modal review generation systems leverage figures, tables, and external knowledge sources to enhance the depth of feedback \citep{taechoyotin2024mamorx}. Reinforcement-learning-enhanced review generators further aim to optimize review comprehensiveness through reward-driven training \citep{taechoyotin2025remor}. Our work builds on these efforts by explicitly incorporating auxiliary context into the review reward signals.

\paragraph{Reinforcement Learning for Text Generation} Reinforcement learning (RL) has been widely applied to natural language generation to enhance training. It requires a clear principled metric that can be applied to a sequence of tokens as a whole rather than token-by-token. Sequence-level training approaches demonstrated that policy gradient methods can directly optimize evaluation metrics rather than token-level likelihoods \citep{ranzato2016sequence}. Actor-critic and self-critical training approaches have since improved stability and sample efficiency in sequence prediction tasks \citep{bahdanau2017actor, rennie2017self}. RL has been successfully applied to tasks such as abstractive summarization and dialogue generation, where reward functions capture semantic coherence, informativeness, or user satisfaction \citep{paulus2017reinforced, li2016deep}. More recently, reinforcement learning from human feedback (RLHF) has become central to aligning large language models with human preferences, enabling instruction-following behavior and improved response quality \citep{christiano2017deep, stiennon2020learning, ouyang2022training}. These advances motivate the use of RL to optimize automated peer reviews toward human-valued criteria such as helpfulness, constructiveness, and factual accuracy.

\paragraph{Group Relative Policy Optimization} Group Relative Policy Optimization (GRPO) has emerged as new reinforcement learning training paradigm designed to improve stability and efficiency in LLM training. GRPO optimizes policies with respect to groups of sequences rather than absolute scalar rewards, thereby reducing variance and improving performance on comparative reasoning \citep{shao2024deepseek, mroueh2025revisiting}. Subsequent work demonstrates its effectiveness in enhancing reasoning capabilities in large language models through structured reward signals \citep{guo2025deepseek}. 

\paragraph{Positioning of the Present Work} While prior automated peer review systems have demonstrated the feasibility of LLM-based feedback generation, several limitations remain. Early automated review generation approaches primarily relied on textual manuscript content alone, which often resulted in shallow critiques and limited factual grounding \citep{bartoli2016paper,kang2018dataset}. More recent LLM-based techniques improve fluency and coverage but still lack sufficient integration of non-textual scholarly signals \citep{lin2023automated,liang2024useful}. Reinforcement learning approaches have introduced reward-driven optimization to improve review helpfulness, reasoning depth, and alignment with human expectations (e.g., \citet{taechoyotin2025remor}). However, these approaches largely operate on textual manuscript content and do not explicitly incentivize models to incorporate figures or external scholarly knowledge sources. Similarly, multi-agent and multi-modal review generation frameworks enhance contextual awareness but typically lack reinforcement-learning-based mechanisms that explicitly reward use of such information \citep{darcy2024marg,taechoyotin2024mamorx}.

The present work extends these lines of research in two key ways. First, we expand reinforcement-learning-based automated peer review generation to incorporate auxiliary contextual information derived from manuscript figures and external knowledge sources. Second, we design reward functions specifically to incentivize the model to use this additional context. These rewards promote contextual awareness with respect to the auxiliary context.

\section{Method}
\label{sec:method}

REM-CTX extends RL-based review generation by introducing \textit{correspondence reward functions} that measure how well a generated review uses the provided auxiliary context. The training objective combines four components: a multi-aspect quality reward, a figure correspondence reward, a novelty correspondence reward, and a formatting reward that encourages structured reasoning traces. We describe each component below.

\subsection{Multi-Aspect Quality Reward}

Review quality is measured using the aspect-coverage framework of \citet{taechoyotin2025remor}, which combines scores across nine dimensions: \textit{criticism}, \textit{example}, \textit{importance \& relevance}, \textit{materials \& methods}, \textit{praise}, \textit{presentation \& reporting}, \textit{results \& discussion}, \textit{suggestion \& solution}, and \textit{METEOR} (manuscript relevance). Each dimension is scored by a fine-tuned classifier that estimates the degree to which a review addresses the corresponding aspect, with scores ranging from 0 to 1. The overall quality reward $\mathcal{R}_{\text{quality}}$ is the sum of all nine dimension scores, yielding a composite metric in $[0, 9]$. We retain identical metrics and classifiers for comparability with prior work.

\subsection{Correspondence Reward Functions}

\begin{figure}[t]
\begin{center}
\includegraphics[width=0.8\textwidth]{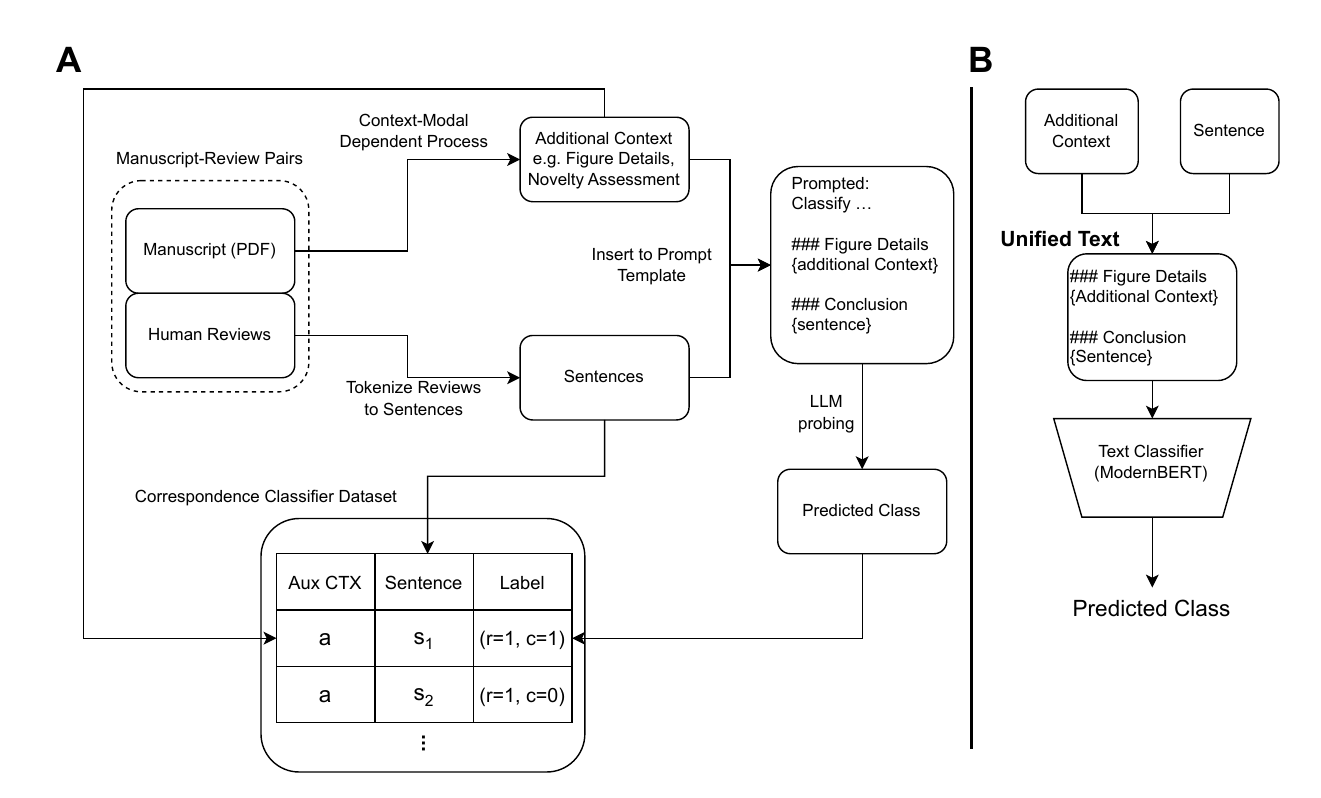}
\end{center}
\caption{Figure Correspondence Reward Function (FCRF) and the Novelty Correspondence Reward Function (NCRF) datasets and model construction. (a)~Sentences from human reviews are paired with auxiliary context (figure details or novelty assessments), and each pair is labeled by an LLM along two axes: relevance and consistency. (b)~A ModernBERT-based classifier is trained on these labels to score new sentence--context pairs.}
\label{fig:correspondence_classifier}
\end{figure}

To promote grounding in auxiliary information, we introduce the Figure Correspondence Reward Function (FCRF) and the Novelty Correspondence Reward Function (NCRF) (Figure \ref{fig:correspondence_classifier}). Both follow the same formulation, differing only in the type of auxiliary context.

\paragraph{Sentence-Level Classification.} For each auxiliary context type, we first obtain an assessment $a$: for FCRF, detailed figure descriptions generated by Sonnet~4. For NCRF, novelty assessments derived from external scholarly data following \citet{taechoyotin2024mamorx}. The NCRF first uses an LLM to generate search keywords based on the title and abstract of the manuscript under review. The generated keywords are used to query a list of similar articles from an external dataset (the Semantic Scholar API). The resulting articles are then used to assess the novelty of each result compared to the manuscript under review. Finally, all the results are summarized by another LLM, and this is the auxiliary context for novelty. After the auxiliary contexts are compiled, we then determine, for each review sentence, whether it is (1) \textit{relevant} to the assessment (i.e., does it discuss the auxiliary context?) and (2) \textit{consistent} with the assessment (i.e., does it agree with the auxiliary content?). See Figure~\ref{fig:correspondence_classifier} (also see Appendices~\ref{app:figure-correspondence-prompt}--\ref{app:novelty-correspondence-prompt}) for the prompts used to generate training labels. This process yields the \textit{FCRDat} and \textit{NCRDat} datasets, which are used to train the downstream classifiers.

\paragraph{Formal Definition.} Let $R = \{s_1, \ldots, s_n\}$ denote the sentences in a generated review and $a$ the auxiliary context. For each sentence $s_i$, we define two binary indicators: relevance $r_i \in \{0, 1\}$ and consistency $c_i \in \{0, 1\}$ (where $c_i{=}1$ denotes non-conflicting). This yields four classes: relevant-consistent ($r{=}1, c{=}1$), relevant-conflicting ($r{=}1, c{=}0$), irrelevant-consistent ($r{=}0, c{=}1$), and irrelevant-conflicting ($r{=}0, c{=}0$).

We train a classifier $f_\theta$ on the ModernBERT \citep{modernbert} embedding of the unified sentence--context text to predict these four classes. The classifier consists of a ModernBERT encoder followed by a linear projection layer with softmax output. The classifier outputs a joint probability:
\begin{equation}
\label{eq:joint_decomp}
    p_\theta(r, c \mid s_i, a) = p_\theta(c \mid r, s_i, a) \cdot p_\theta(r \mid s_i, a)
\end{equation}
For relevant sentences ($r{=}1$), the key quantity is the conditional consistency $p_\theta(c{=}1 \mid r{=}1, s_i, a)$. The FCRF and NCRF achieved a weighted-$F_1$ of 0.69 and 0.66, respectively.

\paragraph{Review-Level Aggregation.} Let $S_{\text{rel}} = \{s_i \in R : \hat{r}_i = 1\}$ be the set of sentences classified as relevant and $S_{\text{con}} = \{s_i \in S_{\text{rel}} : \hat{c}_i = 1\}$ the consistent subset. The correspondence reward is the ratio of consistent sentences among relevant sentences:
\begin{equation}
\label{eq:calc_corresp}
    \text{Corresp}(R, a) = \begin{cases} \dfrac{|S_{\text{con}}|}{|S_{\text{rel}}|} & \text{if } |S_{\text{rel}}| > 0 \\[6pt] 0 & \text{otherwise} \end{cases}
\end{equation}
We instantiate this separately as $\text{Corresp}_{\text{fig}}$ (for figure details) and $\text{Corresp}_{\text{nov}}$ (for novelty assessments). This formulation rewards reviews that, when they discuss auxiliary context, do so accurately. Reviews that never mention auxiliary context receive a reward of zero, providing a gradient signal that encourages the model to engage with the provided context.

\subsection{Training Objective}

The composite reward for a generated review $R$ with auxiliary contexts $a_{\text{fig}}$ and $a_{\text{nov}}$ is:
\begin{equation}
\label{eq:total_reward}
    \mathcal{R}(R) = \mathcal{R}_{\text{quality}}(R) + \text{Corresp}_{\text{fig}}(R, a_{\text{fig}}) + \text{Corresp}_{\text{nov}}(R, a_{\text{nov}}) + \mathcal{R}_{\text{format}}(R)
\end{equation}
where $\mathcal{R}_{\text{format}}$ encourages the model to generate thinking traces (enclosed in \texttt{<think>} tags) prior to the final review. Prior work shows that reasoning traces improve generation quality and structure \citep{shao2024deepseek, guo2025deepseek}. All reward components are weighted equally; we leave exploration of non-uniform weighting to future work. The model is optimized using GRPO \citep{shao2024deepseek} to maximize expected reward across groups of sampled candidate reviews.

\section{Experimental Setup}

\paragraph{Dataset.} We construct \textit{PeerRTEx}, a multi-domain dataset of 234 full-text scientific publications with paired human reviews, collected from the Transparent Peer Review (TPR) initiative, ACL Anthology, and NeurIPS proceedings. The dataset covers three scientific domains: Computer Science ($n{=}130$), Biological Sciences ($n{=}80$), and Physical Sciences ($n{=}24$). Each paper is augmented with figure details obtained via Sonnet~4 and novelty assessments derived from external scholarly data following \citet{taechoyotin2024mamorx} (see above).  For the correspondence classifier training data, we construct \textit{FCRDat} (figure correspondence) and \textit{NCRDat} (novelty correspondence) by pairing segmented review sentences with their respective auxiliary contexts and labeling them using the prompts described in Appendix~\ref{app:figure-correspondence-prompt} and \ref{app:novelty-correspondence-prompt} (see above).

\paragraph{Baselines.} We compare REM-CTX against six systems: (1)~\textit{Vanilla}: simple prompting with Sonnet~4.5; (2)~\textit{Structured}: structured prompting with Sonnet~4.5 \citep{liang2024useful}; (3)~\textit{Multi-Agent}: multi-agent systems \citep{darcy2024marg};  (4)~\textit{MAMORX}: a multi-modal, multi-agent baseline \citep{taechoyotin2024mamorx}; (5)~\textit{Qwen3-8B}: the base model for all RL variants, with simple prompting; and (6)~\textit{REMOR} \citep{taechoyotin2025remor}: RL with multi-aspect quality rewards but no correspondence rewards.

\paragraph{Review Generation Training} Training is conducted on a server with 64 vCPUs, 256 GB RAM, and two A100 (80 GB) GPUs. Reinforcement learning is implemented using TRL \citep{vonwerra2022trl} with Group Relative Policy Optimization (GRPO) \citep{shao2024deepseek}. The effective batch size is 8 (per-device batch size 2 with gradient accumulation 2). Models are trained for 7 epochs, after which performance saturates. Maximum prompt and generation lengths are 32,768 and 4,096 tokens, respectively. The training objective combines the overall review quality reward, correspondence rewards, and thinking trace formats, with uniform weights. Reviews are generated, scored by reward functions, and optimized using GRPO to maximize expected reward.

\section{Results}

\subsection{Overall Comparison}
\begin{figure}[t]
\begin{center}
\includegraphics[width=1\textwidth]{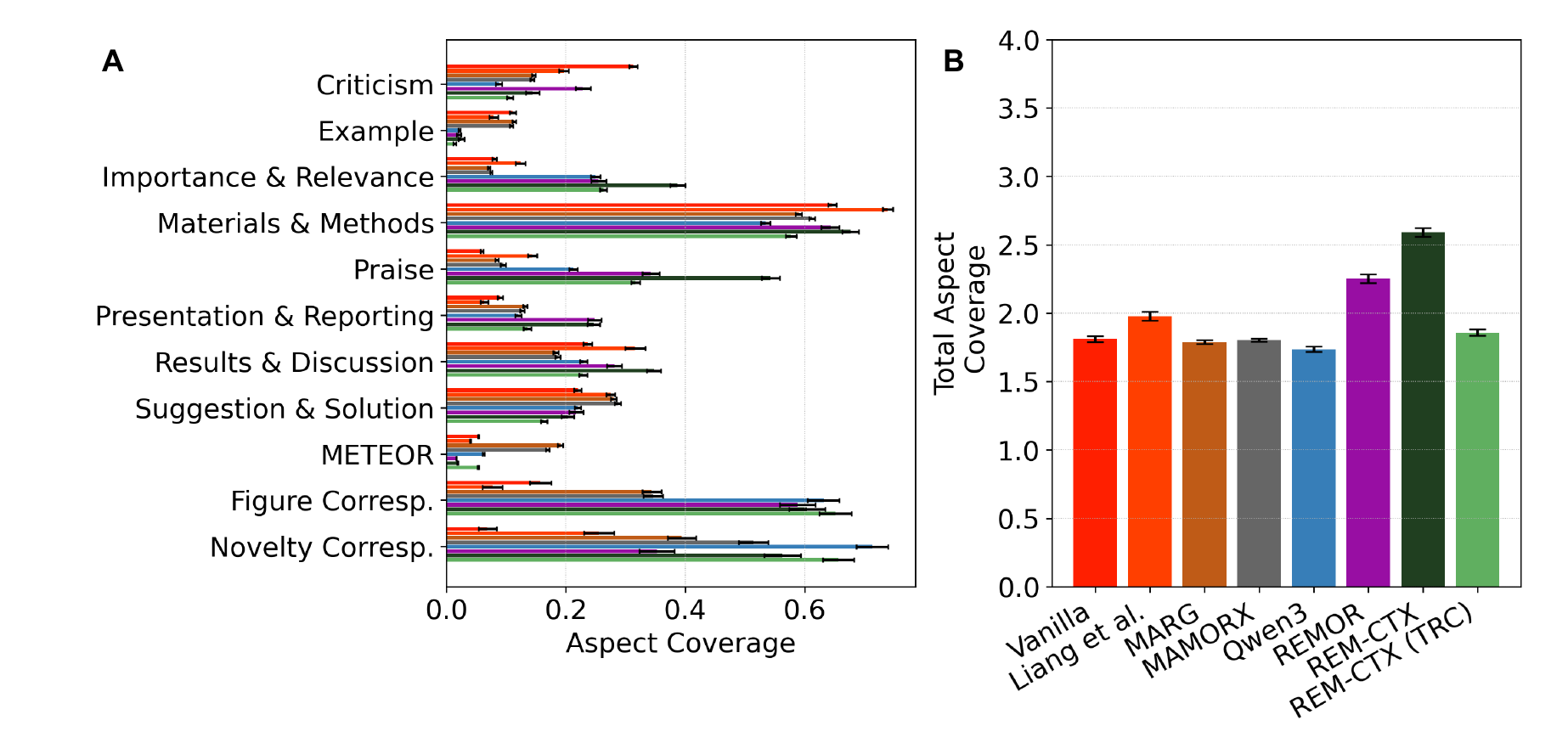}
\end{center}
\caption{The results of Vanilla (Sonnet 4.5), \cite{liang2024useful}, MARG \citep{darcy2024marg}, MAMORX \citep{taechoyotin2024mamorx}, Qwen3-8B, REMOR \citep{taechoyotin2025remor}, and REM-CTX performance across overall review quality (Total Aspect Coverage), dimension coverage, and correspondence reward functions. REM-CTX (TRC) is the score when the thinking traces are included in the evaluation. (a) Dimension and Correspondence scores across models. (b) Overall review quality scores are based on a composite of multiple aspect-specific metrics.}
\label{fig:scores_across_models}
\end{figure}

REM-CTX achieves the highest overall review quality as measured by the total aspect coverage score relative to several baselines (Figure \ref{fig:scores_across_models}B). REM-CTX does not achieve the highest score in all dimensions. The models based on Sonnet 4.5 (Vanilla, Liang et al., MARG, and MAMORX) are more variable, having the highest relative scores in \textit{criticism, example, materials \& methods, suggestion \& solution}, and having the lowest relative scores in \textit{importance \& relevance, praise, presentation \& reporting, results \& discussion} (Figure \ref{fig:scores_across_models}A). This suggests that the underlying model is a strong determinant of the dimensions it prioritizes and de-prioritizes, and that the prompting scheme alone has a more limited influence. 

We found that incorporating auxiliary context improves alignment with provided information but can reduce critical evaluation. The most noticeable pattern is that the model Qwen3-8B achieves a relatively high correspondence score in novelty and figure context, but low scores in several other dimensions. This suggests that the model Qwen3-8B produces reviews that reiterate the provided auxiliary context but may lack criticism. This aligns with the higher METEOR score of Qwen3-8B, which is high when lexical overlap between the review and the manuscript is high, as expected when the model reiterates the provided auxiliary context. 

REMOR has a relatively low novelty correspondence score but a high criticism score, which suggests that the model produces reviews that closely align with the provided novelty assessments but may lack critical evaluation. This pattern is also observed in REM-CTX, which has a higher novelty correspondence score than REMOR but a lower criticism score. This suggests that while REM-CTX produces higher-quality reviews overall, it may generate critiques that conflict with the provided novelty assessments when evaluated in isolation. 

\paragraph{Qualitative Analysis} To better understand the high novelty correspondence score of Qwen3-8B compared to the low score of REM-CTX, we analyze the generated reviews and observe that reviews generated by the base model Qwen3-8B closely mirror the content of the thinking traces of REMOR and REM-CTX and primarily consist of paper summaries accompanied by high-level strengths and weaknesses (see Appendix \ref{app:sample-reviews}). When we included the thinking traces in evaluating aspect coverage (Figure \ref{fig:scores_across_models}), we found that REM-CTX had equivalent or higher correspondence scores than the Qwen3-8B model. This suggests that REM-CTX effectively incorporates the provided auxiliary context into its reasoning process, but chooses to generate more critical and evaluative final reviews that may conflict with the provided novelty assessments when evaluated in isolation. Overall, the behavior of REM-CTX aligns more closely with human reviewers, in which the model first summarizes the paper and then generates more concise critiques. This structure aligns with peer-review guidelines commonly adopted by major publication venues, which encourage reviewers to summarize the work before providing evaluative comments \citep{kelly2014peerreview,plos}.

\begin{figure}[t]
\begin{center}
\includegraphics[width=0.8\textwidth]{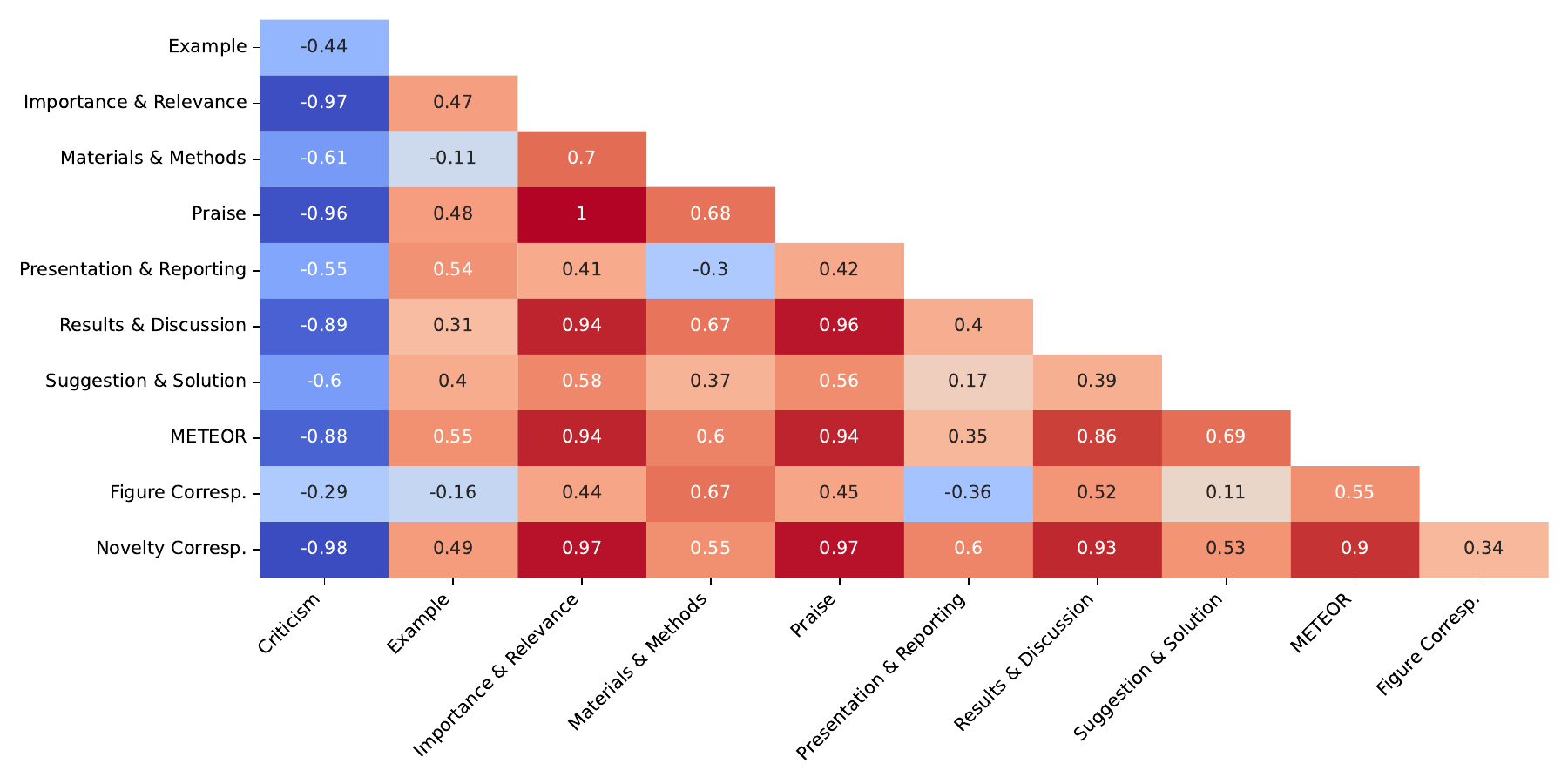}
\end{center}
\caption{Correlation scores of dimension and correspondence scores calculated from a standardized reward value across training epochs (See Appendix \ref{app:std-learning-curve}). This analysis reveals that criticism is negatively correlated with the novelty correspondence score and praise. The presentation \& reporting is negatively correlated with materials \& methods. The presentation \& reporting is negatively correlated with the figure correspondence score.}
\label{fig:scores_across_epochs}
\end{figure}

\paragraph{Trade-off between Aspect Coverage Scores and Correspondence Scores} REM-CTX achieves the highest overall review quality score but has a lower novelty correspondence score than the model Qwen3-8B (Figure \ref{fig:scores_across_models}(a)). This suggests that while REM-CTX produces higher-quality reviews overall, it may generate critiques that conflict with the provided novelty assessments when evaluated in isolation. This pattern is also observed with the model REMOR where it has the highest criticism score but the lowest novelty correspondence score. To expand on the trade-off between dimension scores and correspondence scores, we compare the correlation between each dimension score and the corresponding correspondence score across the training epochs. We find that the novelty correspondence score is negatively correlated with the criticism score (Figure \ref{fig:scores_across_epochs}), implying that the model reduced criticism in favor of better correspondence during training. Furthermore, the results show that the criticism score is negatively correlated with most dimensions, including \textit{importance \& relevance}, \textit{praise}, and \textit{results \& discussion}. It is expected that criticism is negatively correlated with praise. These insights suggest that grouping dimensions could improve model training, a possibility that should be explored in future work.

\begin{figure}[t]
\begin{center}
\includegraphics[width=0.8\textwidth]{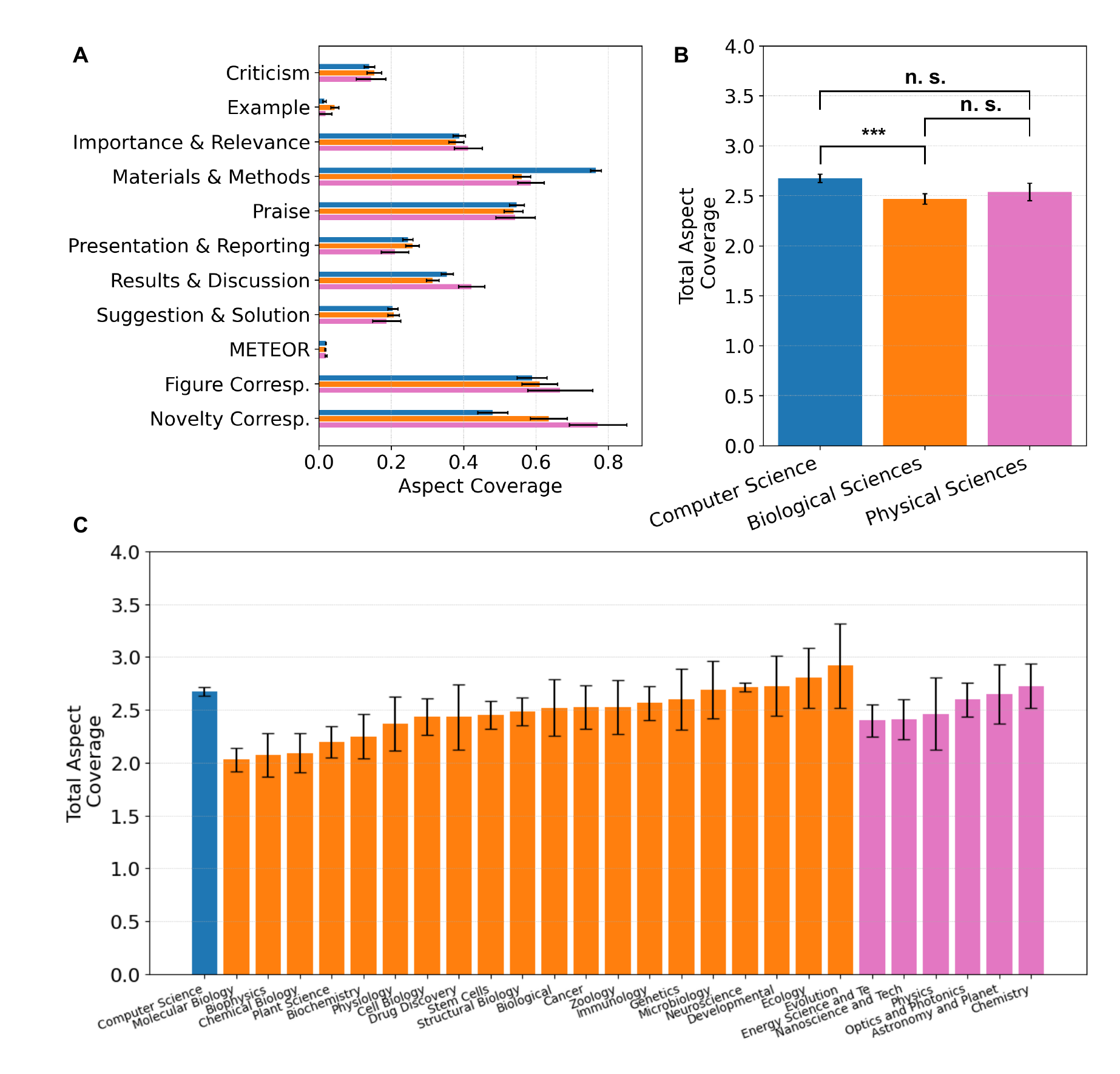}
\end{center}
\caption{REM-CTX scores across Computer Science ($n{=}130$), Biological Science ($n{=}80$), and Physical Science ($n{=}24$). (a)~Per-dimension scores. (b)~Overall quality. Computer Science articles receive significantly higher quality scores than Biological Sciences ($p < 0.01$). (c)~The scores for each minor discipline are within margins of error. Given that all minor disciplines have only 4 papers, except Computer Science, which has 130 papers. Ultimately, this plot suggests that REM-CTX favors all minor disciplines equally.}
\label{fig:scores_by_domain}
\end{figure}

\paragraph{Domain Analysis} We can analyze REM-CTX performance stratified by scientific domain (Figure~\ref{fig:scores_by_domain}). Overall review quality is significantly higher for Computer Science manuscripts than for Biological Sciences ($\text{t}(208) = 3.07, p = .003.$), likely reflecting the composition of the training data, which draws heavily from NeurIPS and ACL proceedings. Physical Sciences manuscripts show intermediate quality scores but with wider variance due to the smaller sample size ($n{=}24$, not-significant).

\paragraph{Ablation Study} To isolate the contribution of each correspondence reward, we compare four REM-CTX variants trained under identical conditions for 7 epochs: \textit{Full} (both correspondence rewards), \textit{Fig-only} (FCRF only), \textit{Novel-only} (NCRF only), and \textit{None} (quality and format rewards only, equivalent to REMOR's reward structure applied to REM-CTX's training setup). The ablation only removes the auxiliary context from the input rather than retraining the model. The ablation yields three findings. Each correspondence reward achieves its targeted objective: figure correspondence is highest when FCRF is included (Full: 0.60, Fig-only: 0.56) and drops when it is not (Novel-only: 0.58, None: 0.54). Symmetrically, novelty correspondence is highest when NCRF is included (Full: 0.56, Novel-only: 0.52) and lower otherwise (Fig-only: 0.52, None: 0.58). This confirms that each reward successfully steers the model toward its intended contextual signal.

\section{Discussion}
In this paper, we aimed to use additional information from a manuscript as auxiliary context in an automated peer review generation system. We introduced two auxiliary correspondence functions to incentivize the model to utilize auxiliary context in reinforcement learning, resulting in higher overall review quality scores and more effective use of the auxiliary context than other similar approaches (e.g., REMOR \cite{taechoyotin2025remor}). We observed that even with a newer model (Sonnet 4.5) using structure prompting or multi-agent systems (e.g., \citet{liang2024useful}), the reviews produced by previous work still cannot match those of a simple prompting scheme.

We found that incorporating thinking traces substantially improves the quality of generated reviews (also see \citet{guo2025deepseek}). Thinking traces closely resemble human reviewers: reviews begin by summarizing the manuscript before articulating targeted critiques (Table \ref{tab:sample_reviews}). The auxiliary rewards make evaluation more context-aware, but it produces a tradeoff with the rest of the  overall review quality metrics (see Figure \ref{fig:scores_across_epochs}). This suggests that grouping dimensions could improve model training, a possibility that should be explored in future work.

\paragraph{Limitations.} The evaluation dataset is modest in size (234 manuscripts), with the Physics domain particularly underrepresented ($n{=}24$). The correspondence classifier's quality directly affects the reward signal; errors can introduce systematic bias, and generalization to out-of-domain content has not been validated. The uniform weighting of reward components (Equation~\ref{eq:total_reward}) is a simplifying choice that may be suboptimal. Finally, automated peer review systems can exhibit biases, including affiliation bias \citep{vonwedel2024affiliation}, and the present work does not include a bias audit.

\section{Conclusion}
We introduced REM-CTX, an RL-based framework for automated peer review that incorporates auxiliary context through correspondence reward functions. On a dataset of 234 manuscripts spanning three scientific domains, REM-CTX achieves the highest overall review quality among six baselines. Ablation studies confirm that the two correspondence rewards are complementary, each targeting its intended dimension without degrading others, and that their combination yields the best overall performance. Analysis of training dynamics reveals interpretable trade-offs between criticism and contextual correspondence, connecting to broader challenges in multi-objective reward optimization for language models.  Future work may extend the introduction of auxiliary context idea to citations, tables, and other scholarly signals, and explore adaptive reward weighting. We release the \textit{PeerRTEx}, \textit{FCRDat}, and \textit{NCRDat} datasets to support further research.

\section*{Ethics Statement}

Automated peer review systems are intended as assistive tools to support, not replace, human reviewers. We recognize that such systems may propagate biases present in training data or exhibit systematic biases such as affiliation bias \citep{vonwedel2024affiliation}. We advocate for transparency in the use of automated review tools and emphasize the importance of human oversight in all peer review decisions. The datasets used in this work are derived from publicly available peer review records.

\bibliographystyle{unsrt}  
\bibliography{references}  

\appendix
\section{Prompts}
To obtain the FCRDat and NCRDat, we used the following prompt templates to probe a commercial large language model to get the result. We can see that the prompt is organized as a unified text, similar to the template for the correspondence classifier.

\subsection{Figure Correspondence Prompt}
\label{app:figure-correspondence-prompt}
For the figure auxiliary context, the \textit{figure detail-sentence} pairs are classified into four classes. The four classes are derived from the joint binary indicators of relevance and consistency. Relevance is defined as the mention of figures, and consistency is adherence to the provided figure details.

\begin{lstlisting}
Please classify whether the following premise and conclusion are relevant to each other or not. Answer only 0, 1, 2 or 3. Do not include more details.
Answer 0 for when the conclusion involves the figure and the conclusion is drawn from the figure details
Answer 1 for when the conclusion involves the figure but the conclusion conflicts with the figure details
Answer 2 for when the conclusion does not involve the figures and the conclusion is drawn from the figure details
Answer 3 for when the conclusion does not involve the figures but the conclusion conflicts with the figure details

### Figure Details
{figure details}

### Conclusion
{sentence of interest}

\end{lstlisting}

\subsection{Novelty Correspondence Prompt}
\label{app:novelty-correspondence-prompt}
For the novelty auxiliary context, the \textit{novelty assessment-sentence} pairs are classified into four classes. The four classes are derived from the joint binary indicators of relevance and consistency. Relevance is defined as the comments commenting on the novelty of the work, and consistency is defined as being consistent with the provided novelty assessment.

\begin{lstlisting}
Please classify whether the following premise and conclusion are relevant to each other or not. Answer only 0, 1, 2 or 3. Do not include more details.
Answer 0 for when the conclusion concerns novelty and the conclusion is drawn from the novelty assessment
Answer 1 for when the conclusion concerns novelty but the conclusion conflicts with the novelty assessment
Answer 2 for when the conclusion does not concern novelty and the conclusion is drawn from the novelty assessment
Answer 3 for when the conclusion does not concern novelty but the conclusion conflicts with the novelty assessment

### Novelty Assessment
{novelty assessment}

### Conclusion
{sentence of interest}
  
\end{lstlisting}

\section{Additional Results}
\subsection{Standardized Learning Curve}
\label{app:std-learning-curve}
\begin{figure}[t]
\begin{center}
\includegraphics[width=1\textwidth]{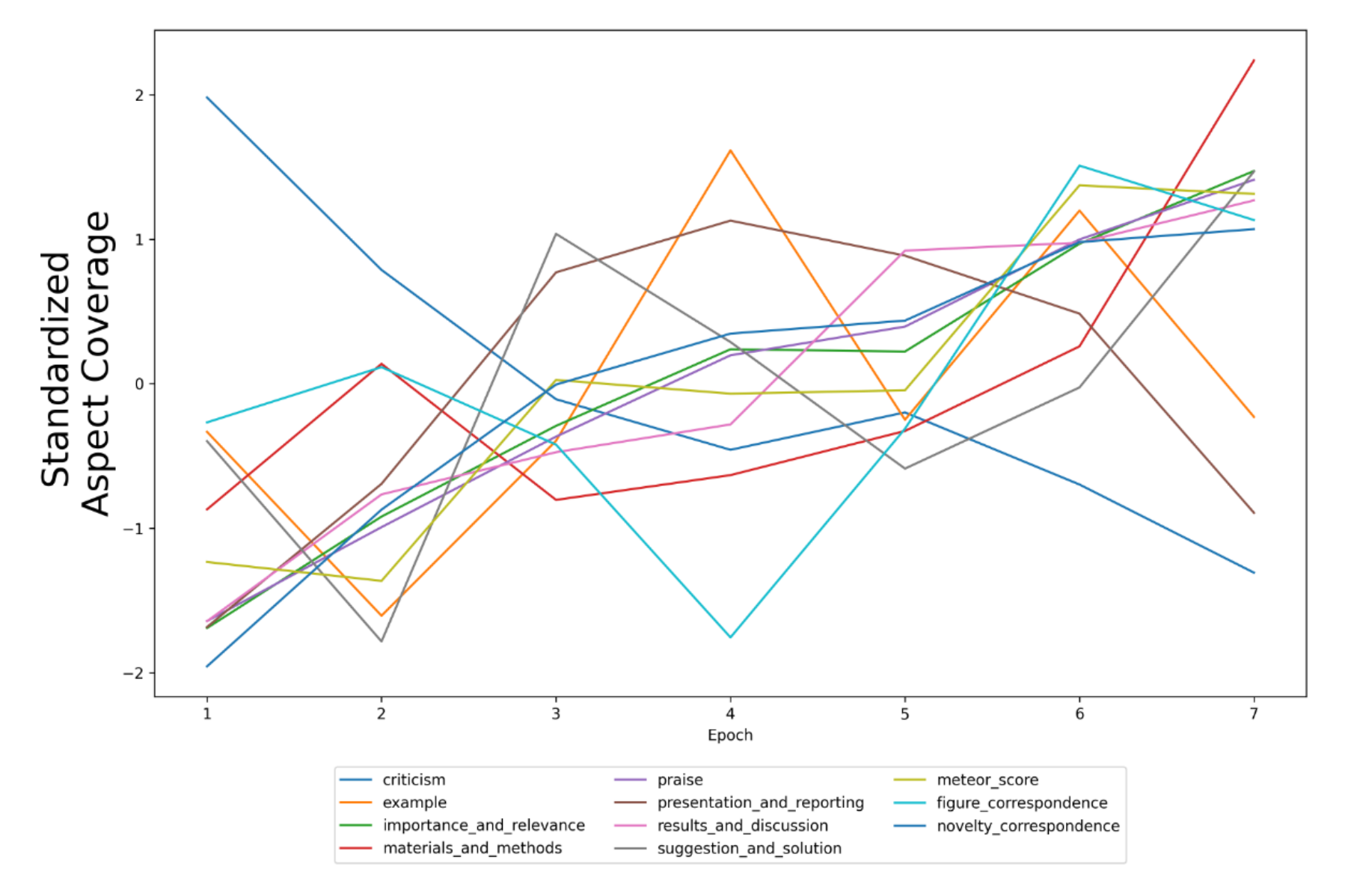}
\end{center}
\caption{The plot shows a standardized learning curve for each reward dimension. It shows that \textit{criticism} is the only dimension that decreases over time. This is an interesting decision for the model to prioritize other dimensions at the cost of \textit{criticism}.}
\label{fig:scores_across_epochs_std}
\end{figure}

To understand the optimization priority of the training process, we standardized each metric's scores per epoch (Figure \ref{fig:scores_across_epochs_std}). The standardization per metric accounts for the inherent difficulty of optimizing each metric and makes the metrics comparable. With this result, the relationship between metrics becomes clear. The results show that some metrics, such as \textit{criticism}, decline with each epoch, while \textit{materials \& methods} increases. We can see that \textit{METEOR} is increasing as well, given that the absolute score is barely a tenth of other metrics.

\subsection{Review Comparison between Qwen3-8B and REM-CTX}
\label{app:sample-reviews}
To understand the results beyond quantitative measures, we analyzed the reviews produced by each model. At a high level, the reviews generated look similar, but a deep dive reveals that those produced by \textit{REM-CTX} are more concise and to the point than \textit{Qwen3-8B} (Table \ref{tab:sample_reviews}). This conciseness is crucial for the authors receiving these reviews because it helps them focus on what needs to be addressed and what they did well.

\begin{table}[t]
\begin{center}
    \scriptsize
    \begin{tabular}{p{0.40\linewidth}|p{0.40\linewidth}}
        \toprule
        \textbf{Qwen3-8B}  & \textbf{REM-CTX}\\
        \midrule
                \begin{lstlisting}
### Peer Review Summary

**Overall Evaluation:**
The paper presents a novel approach to multi-modal sentiment analysis by incorporating contextual relationships among utterances using LSTM networks. ...
---
### **Strengths:**

1. **Comprehensive Literature Review**: The paper provides a thorough overview of existing work ...

2. **Clear Methodology**: The proposed method is well-explained, with a structured approach to extracting both unimodal and multi-modal features...
        \end{lstlisting}
 & \begin{lstlisting}
<think>
...
3. **Strengths**:
   - The paper addresses an important problem in sentiment analysis (modeling contextual relationships)...

   Overall, ...
</think>
The authors propose an LSTM network that models the contextual relationships between utterances in a video for sentiment analysis. The paper is well-written and the proposed model is interesting, but the experimental results are not convincing enough to recommend acceptance. The authors compare their model with several baseline methods, but it is not clear whether these baselines are state-of-the-art methods or not...
\end{lstlisting} \\
        \bottomrule
    \end{tabular}
\end{center}
\caption{The review generated by Qwen3-8B closely mirrors the content of the thinking traces of REM-CTX and primarily consists of paper summaries accompanied by high-level strengths and weaknesses.}\label{tab:sample_reviews}
\end{table}

\end{document}